\documentclass{article}
\usepackage{svg}
\usepackage{ragged2e}
\usepackage{booktabs}
\usepackage{array}
\usepackage{amsfonts}
\usepackage{amsmath}
\usepackage{stfloats} %
\usepackage[preprint]{corl_2026} 

\title{FT-WBC: Learning Fault-Tolerant Whole-Body Control for Legged Loco-Manipulation}

\author{
  \textbf{Yudong Zhong}\textsuperscript{1},
  \textbf{Pengfei Mai}\textsuperscript{1},
  \textbf{Sikai Guo}\textsuperscript{1},
  \textbf{Jiahang Cao}\textsuperscript{2},
  \textbf{Zhihai Bi}\textsuperscript{1}\\
  \textbf{Qiuyue Liu}\textsuperscript{1},
  \textbf{Ziyan Feng}\textsuperscript{1},
  \textbf{Jinni Zhou}\textsuperscript{1},
  \textbf{Jun Ma}\textsuperscript{1}\\
  \textsuperscript{1}The Hong Kong University of Science and Technology (Guangzhou)\\
  \textsuperscript{2}The University of Hong Kong\\
  \url{https://ft-wbc.github.io/}
}
\usepackage{multirow}
\usepackage{graphicx}
\usepackage{svg}
\usepackage{subcaption}
\usepackage{wrapfig}
\usepackage{float}
\usepackage[utf8]{inputenc}
\usepackage{caption}
\begin{document}
\maketitle

\begin{figure*}[ht]
    \centering
    \vspace*{-2.0em}
    \includegraphics[width=.95\linewidth]{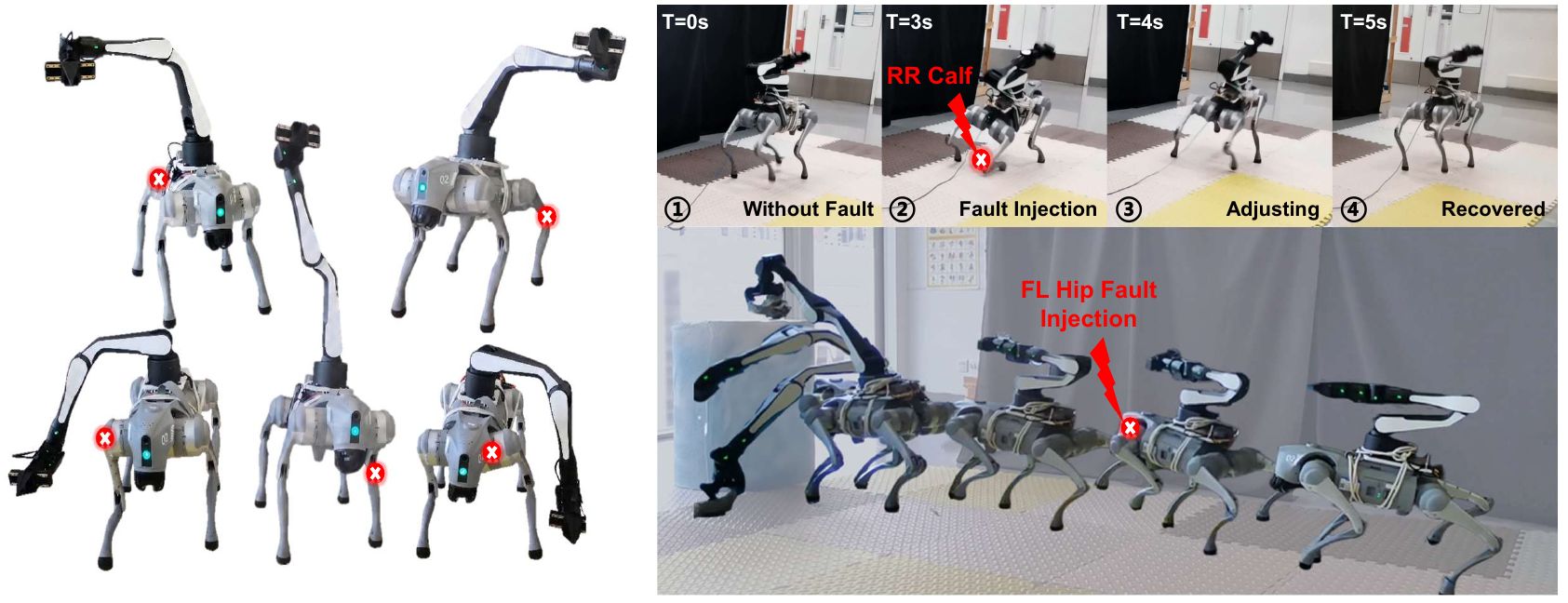}
    \vspace{-1mm}
    \caption{\justifying We present FT-WBC, a framework that enables robust, fault-tolerant whole-body control for legged loco-manipulation under actuator failure. FT-WBC governs an arm-mounted quadruped manipulator to dynamically adapt its whole-body posture and synthesized compensatory gaits upon detecting severe actuator failures. The left panel illustrates the whole-body control capability of our framework under failures in different joints. The upper-right panel shows the robot's rapid gait recovery when the rear right calf joint loses actuation during walking, while the lower-right panel demonstrates a complete reaching and pick-and-place task under a front left hip joint failure.}
    \vspace{-3mm}
    \label{fig:teaser_figure}
\end{figure*}

\begin{abstract}
    Legged manipulators combine the mobility of legged platforms with the manipulation capability of robotic arms. However, arm-induced Center-of-Mass shifts and dynamic disturbances make the system more prone to instability under actuator failures, potentially leading to falls, task failures, or safety risks. Existing fault-tolerant control methods mainly focus on locomotion alone, leaving the coupled problem of whole-body stability and arm reachability in fault-tolerant loco-manipulation largely unaddressed. To bridge this gap, we propose FT-WBC, a fault-tolerant loco-manipulation framework for robust whole-body control of legged manipulators under actuator failures. FT-WBC adopts a decoupled upper- and lower-body policy architecture and introduces two key modules: a Fault Estimator (FE) and a Posture Adaptation Module (PAM). The FE predicts faulty joints from lower-body proprioceptive histories, while the PAM uses this fault information to adapt the base posture plan generated by the arm policy, converting potentially unstable posture requests into safe and executable base posture commands. Through this fault-aware posture adaptation mechanism, FT-WBC synthesizes compensatory gaits under actuator failures and preserves as much arm workspace as possible while maintaining whole-body stability. Simulation and real-world experiments show that FT-WBC significantly improves survival rate and workspace under weakening or locked failures, and transfers zero-shot to a real legged manipulator in the real world. 
\end{abstract}



\section{Introduction}
\label{sec:Introduction}

Integrating a robotic arm to a legged platform is a prevalent configuration of mobile manipulation systems. In such systems, the robot not only needs to move close to the target, but also uses body pitch or roll to assist the robotic arm in performing manipulation tasks. Such whole-body manipulation capabilities allow robots to perform more flexible reaching and interaction tasks~\cite{LM2, LM11, LM15}, but they also introduce greater risks of actuator failures during real-world deployment. Unlike locomotion alone, loco-manipulation often requires the robot to maintain tilted base postures for extended periods to expand the arm workspace, which can place sustained loads and torques on specific lower-limb joints. These concentrated loads increase the risk of motor overheating or joint degradation. Such failures can lead to loss of stability, object drops, collisions, or even safety risks to nearby humans. Therefore, legged loco-manipulation systems intended for real-world deployment must be able to continue executing tasks safely under actuator failures.

Existing studies on fault-tolerant control mainly focus on quadrupedal locomotion. Recent methods based on deep reinforcement learning (DRL) have improved the ability of legged robots to maintain locomotion under diverse fault conditions~\cite{F1, F2, F3}. However, these methods are primarily designed for robots without an active manipulator and mainly target fall prevention and velocity-command tracking. Fault-tolerant loco-manipulation introduces a more demanding stability problem: the mounted arm raises the Center-of-Mass (CoM) and introduces additional dynamic disturbances, making balance harder to maintain when lower-limb actuators fail. Meanwhile, preserving an effective arm workspace often requires base pitch or roll, which may become unsafe after a failure.

To address this gap, this paper proposes FT-WBC, a fault-tolerant whole-body-control framework that balances locomotion stability with workspace expansion under actuator failures. FT-WBC consists of two key modules: the Fault Estimator (FE) and the Posture Adaptation Module (PAM). The FE estimates faulty joint conditions from historical lower-body observations, providing fault information for robust locomotion. The PAM uses this information to adapt manipulation-driven base posture commands, preserving arm reachability while maintaining base stability under joint failures. Our main contributions are summarized as follows:
\begin{enumerate}
    \item We propose FT-WBC, a fault-tolerant loco-manipulation framework for legged manipulators. The framework coordinates lower-body stability control with upper-body manipulation requirements under joint failures, allowing the robot to maintain robustness while executing whole-body loco-manipulation tasks;
    \item We introduce two key modules for robust fault-tolerant control: the FE and the PAM. The FE predicts faulty joints and provides a fault vector for adaptive compensation of lower-limb failures. The PAM utilizes this fault information to revise unstable posture commands, thereby improving base stability and arm workspace under faults;
    \item We achieve zero-shot sim-to-real transfer of FT-WBC. Extensive simulation and real-world experiments demonstrate that the framework enables robust control under diverse joint failures and successfully completes object grasping and placing tasks.
\end{enumerate}


\section{Related Work}
\label{sec:Related Work}

    \subsection{Legged Loco-Manipulation}
    \label{subsec:Legged Loco-Manipulation}
    Legged loco-manipulation research first focuses on low-level whole-body control, where the main objective is to coordinate the legs and the arm so that the robot can adjust its base posture to expand the reachable workspace of the manipulator~\cite{LM2,LM5,LM14,LM13}. Building on this foundation, subsequent studies further enhance the physical interaction capability of low-level controllers, including contact force regulation, heavy-payload transport, and more stable whole-body coordination~\cite{LM9,LM6,LM8}, allowing legged manipulators to remain effective in more complex object-interaction tasks.

    For more complex interaction tasks, relying on the arm alone is not always sufficient. This motivates studies that explore tighter coordination between the legs and the arm, such as coupling upper-limb manipulation with lower-limb motions to accomplish complex tasks, or mounting small manipulators on the front legs of a quadruped so that the legs can also participate in manipulation~\cite{LM3,LM12}. Other work uses the robotic arm as an auxiliary balance mechanism to improve stability during dynamic motion~\cite{LM10}. Meanwhile, stable low-level control alone is still insufficient for autonomous mobile manipulation in open environments.

    Therefore, recent work further introduces high-level planning, visual or language instructions, and more complex leg-arm coordination strategies to improve task autonomy and generalization. For example, visual and linguistic information has been used to guide robots through diverse tasks~\cite{LM11,LM1,LM15,LM7,LM16,LM17}, and some studies investigate posture optimization under constrained viewpoints to improve perception and manipulation~\cite{LM4}. These works expand the capability boundary of legged loco-manipulation from low-level control, system morphology, and high-level task planning.

    However, these capabilities mostly rely on the assumption of nominal actuator operation. Once actuator failures occur, the robot may no longer reliably generate the required support forces or execute base posture adjustments, making existing loco-manipulation capabilities unstable or even unsafe. Therefore, achieving fault-tolerant whole-body control under actuator failures remains an underexplored yet important problem in legged loco-manipulation.

    \subsection{Fault-Tolerant Locomotion with DRL}
    \label{subsec:fault_tolerant_locomotion}
    Existing fault-tolerant locomotion studies have substantially improved the ability of quadruped robots to keep moving under joint failures. Early work mainly focuses on recovery after sudden failures, enabling robots to regain stable locomotion after actuator damage~\cite{F4,F6,F9,F10,F13,F14}. Yet, these methods often consider a limited set of fault conditions. To improve coverage, subsequent studies consider different faulty joints, fault severities, and failure modes, allowing robots to maintain locomotion under a broader range of predefined fault settings~\cite{F5,F7,F3,F12,F15}. Since predefined robustness does not fully address failures outside the training distribution, more recent methods improve adaptation to unseen failures and reduce reliance on training-specific fault patterns~\cite{F1,F11}, thereby enhancing overall system operational resilience.

    Nevertheless, these methods still focus on locomotion alone, such as maintaining balance, recovering gait, and tracking velocity commands. For a legged manipulator, the arm introduces additional payload and CoM shifts, and actively requires base posture adjustments to expand the end-effector workspace. As a result, actuator failures affect not only whether the robot can keep moving, but also whether the arm can perform manipulation stably and effectively. Existing locomotion-oriented fault-tolerant control methods are therefore difficult to directly apply to this coupled stability-and-reachability problem. To address this gap, we study fault-tolerant loco-manipulation and propose FT-WBC, which coordinates lower-body stability control and upper-body manipulation requirements in a fault-aware manner.

\section{Methods}
\label{sec:Methods}

\subsection{Framework Overview}
\begin{figure}
    \centering
    \includegraphics[width=1\linewidth]{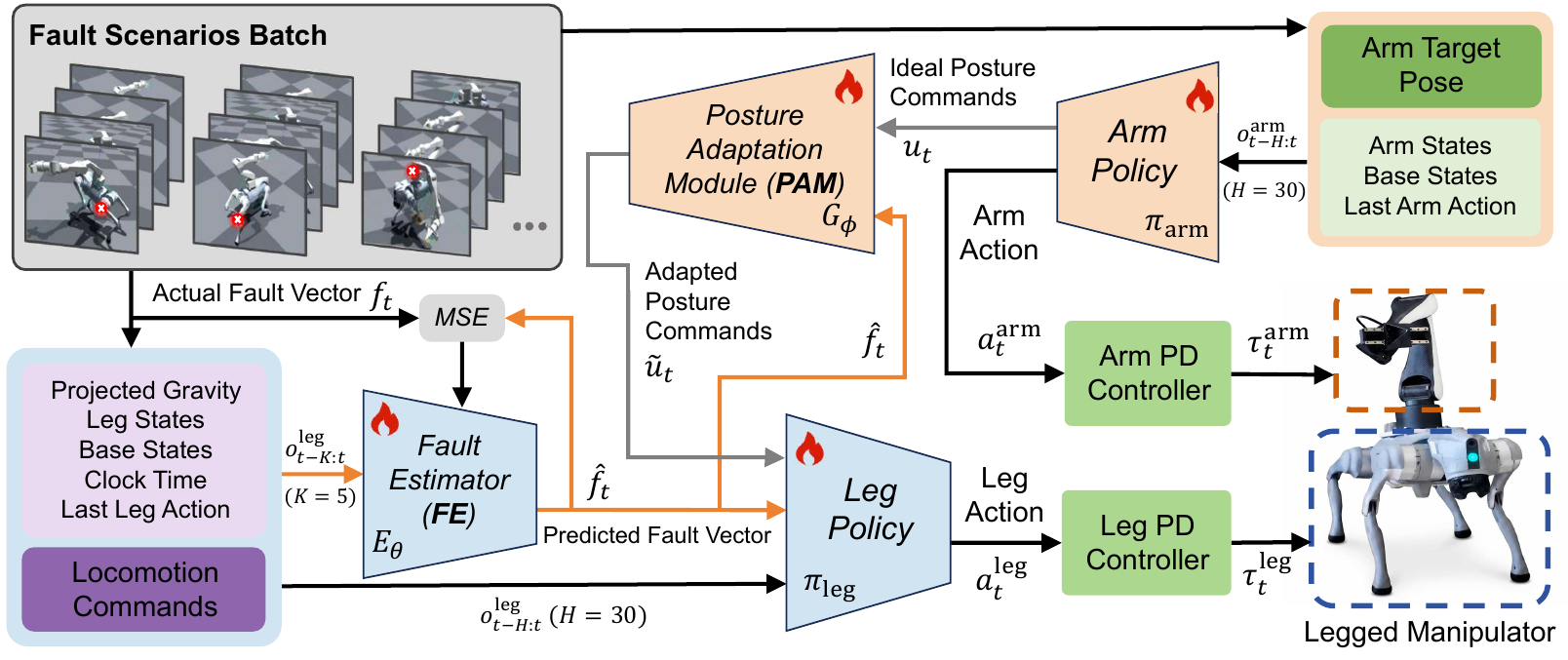}
    \caption{\justifying Overview of FT-WBC. The framework adopts a decoupled architecture consisting of a lower-body leg policy and an upper-body arm policy. The leg policy generates quadruped joint actions from target motion commands and the predicted fault vector, while the arm policy tracks the target end-effector pose and outputs both arm joint actions and a desired base posture plan. The FE predicts joint fault conditions, and the PAM remaps the desired base posture plan into a safe posture command, enabling fault-aware whole-body control under actuator failures.}
    \label{fig:framework}
\end{figure}

As illustrated in Figure \ref{fig:framework}, FT-WBC adopts a fault-aware decoupled upper- and lower-body framework that connects upper-body manipulation intent, fault estimation, posture adaptation, and lower-body stability control into a closed-loop pipeline. Specifically, the arm policy $\pi_{\mathrm{arm}}$ takes the arm observation history $\boldsymbol{o}_{t-H:t}^{\mathrm{arm}}$ as input and generates the arm action $\boldsymbol{a}_t^{\mathrm{arm}}$ together with a desired base posture plan $\boldsymbol{u}_t$, where the target end-effector pose command is denoted by $\boldsymbol{c}_t^{\mathrm{arm}}$. To support stable control under joint failures, FE $E_{\theta}$ estimates the current joint-fault condition from the lower-body proprioceptive history $\boldsymbol{o}_{t-H:t}^{\mathrm{leg}}$ and outputs the predicted fault vector $\hat{\boldsymbol{f}}_t$, which provides explicit fault information to the leg policy $\pi_{\mathrm{leg}}$. The PAM $G_{\phi}$ then uses the fault information from $E_{\theta}$ to safely remap the base posture plan $\boldsymbol{u}_t$ into an adapted posture command $\tilde{\boldsymbol{u}}_t$, preventing posture commands that may shift the CoM toward a degraded support side under faults. Finally, $\tilde{\boldsymbol{u}}_t$, $\hat{\boldsymbol{f}}_t$, and $\boldsymbol{o}_{t-H:t}^{\mathrm{leg}}$ are fed into $\pi_{\mathrm{leg}}$, allowing it to generate lower-body joint actions $\boldsymbol{a}_t^{\mathrm{loco}}$ for compensatory gait synthesis and stable whole-body loco-manipulation under joint failures. Detailed observations and actions of $\pi_{\mathrm{arm}}$ and $\pi_{\mathrm{leg}}$ are provided in Appendix~\ref{app:leg_arm_policy}.
     
\subsection{Fault Estimator}
\label{subsec:fault_estimator}
Merely injecting faults during training falls far short of achieving the requisite robustness. In real-world physical interactions, actuator faults are quintessentially hidden states. The original Actor network struggles to formulate effective responses relying solely on shallow features.
    
Therefore, we introduce the FE to decouple fault information from the continuous stream of proprioception. This module utilizes a lower-body observation history $\boldsymbol{o}_{t-K:t}^{\mathrm{leg}}$ of length $K=5$ as input, aggregating residual information across the temporal dimension to evaluate the dynamic responsiveness of the actuators. The input tensor is reconstructed and fed into a Multi-Layer Perceptron (MLP). Here, $\text{MLP}_{\text{enc}}$ serves as a high-capacity mapper, extracting a fault-sensitive latent $\boldsymbol{z}_t$:
$$\boldsymbol{z}_t = \text{MLP}_{\text{enc}}(\boldsymbol{o}_{t-K:t}^\mathrm{leg}), \quad \hat{\boldsymbol{f}}_t = \sigma(\boldsymbol{W}_f \boldsymbol{z}_t + \boldsymbol{b}_f), $$
where $\sigma$ denotes the Sigmoid activation function, and $\boldsymbol{W}_f$ and $\boldsymbol{b}_f$ represent the learnable weights and biases of the projection layer. This yields a continuous predicted fault vector $\hat{\boldsymbol{f}}_t \in \mathbb{R}^{12}$, where each element $\hat{f}_j \in [0,1]$ indicates the predicted fault probability of the corresponding leg joint.
    
We use a binary fault vector $\boldsymbol{f}_t \in \mathbb{R}^{12}$ as the supervision signal for the FE, where each element $f_j$ corresponds to one leg joint, with $f_j=1$ indicating a fault and $f_j=0$ indicating a healthy joint. The FE outputs a predicted fault vector of the same dimension and is trained with a Mean Squared Error (MSE) loss to minimize the difference between the prediction and the ground-truth label:
$$\mathcal{L}_{\text{fault}} = \frac{1}{12} \sum_{j=1}^{12} (\hat{f}_j - f_j)^2.$$
    
\textbf{Warm-up Mechanism:} To prevent the predictor's initial instability (random guessing) from degrading the RL policy's learning at the early stages, the Actor uses ground-truth fault labels for the first $N = 3000$ iterations, before seamlessly switching to the outputs generated by the FE.
   
\subsection{Posture Adaptation Module}
\label{subsec:posture_adaptation}
In loco-manipulation, arm reachability conflicts with base stability. For example, grasping a front-right object requires pitching the base forward-right. If the front-right leg fails and the base blindly executes this posture plan, the CoM shifts directly onto the incapacitated leg. Lacking ground reaction force, a catastrophic fall is inevitable.
    
To address this conflict, we introduce the PAM. In the system hierarchy, this module acts as a high-level intent arbitrator. It deeply fuses the upper-body's original posture plan $\boldsymbol{u}_t$ and the predicted fault vector $\hat{\boldsymbol{f}}_t$ in the feature dimension, feeding them into a non-linear mapping network for intent reconstruction. PAM maps the $\pi_{\mathrm{arm}}$ posture plan into a safe posture-command space:
$$\tilde{\boldsymbol{u}}_t = \tanh(\boldsymbol{W}_o \boldsymbol{h}_t + \boldsymbol{b}_o),$$
where $\boldsymbol{h}_t$ represents the fused hidden feature representation, while $\boldsymbol{W}_o$ and $\boldsymbol{b}_o$ are the learnable weights and biases of the output layer.

The normalized output is then scaled and clipped within the robot's physical boundaries:
$$\tilde{c}_t^p = \text{clip}(0.4 \tilde{u}_t^p, -0.3, 0.3), \quad \tilde{c}_t^r = \text{clip}(0.4 \tilde{u}_t^r, -0.2, 0.2), $$
where $\tilde{u}_t^p$ and $\tilde{u}_t^r$ are the pitch and roll components of $\tilde{\boldsymbol{u}}_t$. Thus, $\tilde{c}_t^p$ and $\tilde{c}_t^r$ are clipped to $[-0.3, 0.3]$ and $[-0.2, 0.2]$ radians, respectively, preventing infeasible base postures before command execution.
        

\section{Experiment Results}
\label{sec:result}

To evaluate the effectiveness and robustness of our proposed fault-tolerant whole-body control framework, we conduct extensive experiments in both simulation and the real world. 

\textbf{Experimental setup:} Our hardware platform consists of a Unitree Go2 quadruped equipped with an Airbot Play robotic arm. The control frequency for both the upper and lower body is strictly set to $50\,\mathrm{Hz}$ across all environments. For policy training, we utilize the Isaac Gym simulator powered by an NVIDIA RTX 4090 GPU. The training process lasts for a total of 80,000 PPO iterations. To stabilize the initial learning phase, we employ a curriculum strategy: during the first 10,000 iterations, the robotic arm is fixed to pre-train the base locomotion capability; subsequently, the full whole-body loco-manipulation training is activated. For real-world deployment, the learned policy is deployed zero-shot on a Jetson Orin Nano ($8\,\mathrm{GB}$) onboard computer for real-time inference.

\textbf{Baselines and ablations:} We use Robo-Duet \cite{LM14} as our primary baseline, representing a state-of-the-art decoupled loco-manipulation framework without explicit fault-tolerance mechanisms. We further compare our method with two ablated variants: without the FE (\textbf{w/o FE}) and without the PAM (\textbf{w/o PAM}). In \textbf{w/o FE}, the PAM uses ground-truth fault labels instead of FE predictions.

\subsection{Evaluation of the FE and PAM}
\label{subsec:fault_prediction_eval}
        
\textbf{Fault localization:} The effectiveness of our adaptive whole-body control hinges on the system's ability to precisely localize which actuator is impaired. To evaluate the performance of the FE ($E_\theta$), we focus on its capability to identify the specific faulty joint index among the 12 leg actuators.

\begin{figure}[htbp]
    \centering
    \includegraphics[width=1\linewidth]{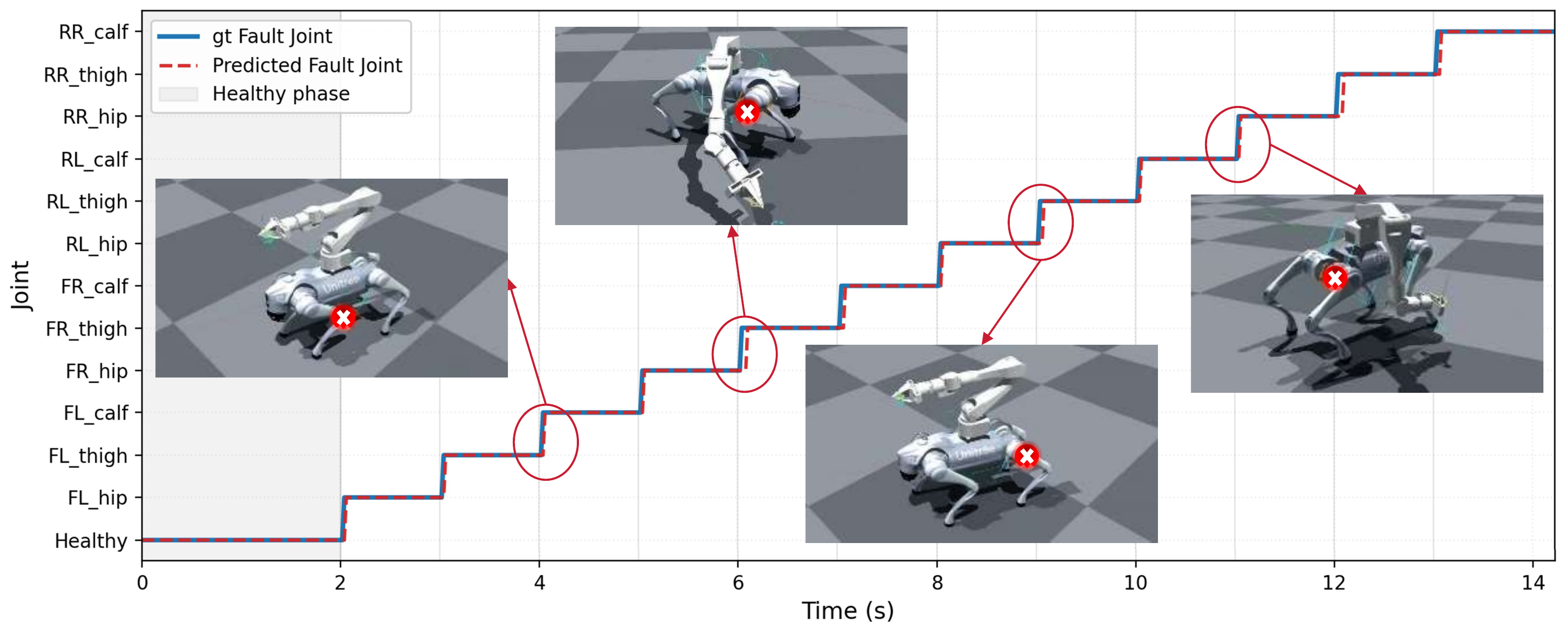}
    \caption{Evaluation of the FE. The x-axis denotes time, and the y-axis denotes the joint name. Different joint faults are injected every $1\,\mathrm{s}$. The solid blue line indicates the ground-truth fault injection, while the dashed red line represents the predicted fault index. GT denotes ground truth.}
    \label{fig:fault_evaluation}
\end{figure}

As illustrated in Figure \ref{fig:fault_evaluation}, we present a timeline of fault identification. The x-axis represents time, and the y-axis represents the joint name, with different joint failures injected every $1\,\mathrm{s}$. At $t = 2.0\,\mathrm{s}$, an initial  failure is injected. The predicted fault index closely follows the ground-truth fault over time. Despite noisy proprioceptive data, the module filters transient oscillations and locks onto the correct joint index within approximately $0.05\,\mathrm{s}$ after injection. 

\textbf{Posture adaptation:} We further validate the effectiveness of the PAM by testing two representative faulty joints on the FL leg and two end-effector targets in the FL lower region. Reaching such targets normally induces leftward roll and forward/downward pitch, which can shift the body load toward the impaired FL support. PAM suppresses these risky fault-directed posture adjustments while pre-

\begin{wraptable}{r}{0.57\textwidth}
    \vspace{-0.8em}
    \centering
    \scriptsize 
    
    \setlength{\tabcolsep}{6pt} 
    
    \captionsetup{font=scriptsize, justification=centering}
    \caption{Evaluation of PAM.}
    \label{tab:fl_posture_commands}
    
    \begin{minipage}{0.55\textwidth}
        \centering
        \begin{tabular}{ccccc}
        \toprule
        \textbf{Fault} & \textbf{Method} & \textbf{Target LPY} & $\boldsymbol{R_{\mathrm{FL}}}\downarrow$ & $\boldsymbol{S_{\mathrm{FL}}}\downarrow$ \\
        \midrule
        \multirow{3}{*}{\begin{tabular}{@{}c@{}}FL Thigh\\$k_\tau=0.00$\end{tabular}}
         & RD & \multirow{3}{*}{\begin{tabular}{@{}c@{}}$(0.607,$\\$0.089,$\\$0.720)$\end{tabular}} & $41.7\%$ & $0.0706$ \\
         & w/o PAM &  & $41.4\%$ & $0.0680$ \\
        & \textbf{Ours} &  & $\mathbf{39.2\%}$ & $\mathbf{0.0000}$ \\
        \midrule
        \multirow{3}{*}{\begin{tabular}{@{}c@{}}FL Calf\\$k_\tau=0.25$\end{tabular}}
         & RD & \multirow{3}{*}{\begin{tabular}{@{}c@{}}$(0.520,$\\$0.403,$\\$0.316)$\end{tabular}} & $30.7\%$ & $0.1672$ \\
        & w/o PAM &  & $30.3\%$ & $0.1640$ \\
        & \textbf{Ours} &  & $\mathbf{20.0\%}$ & $\mathbf{0.0000}$ \\
        \bottomrule
        \end{tabular}
        \vspace{0.4em}
        
        \par
        \begingroup
        \captionsetup{justification=justified, singlelinecheck=off}
        \noindent\textit{Note:} Target LPY is $(l,p,y)$, with $l$ in meters and $p,y$ in radians. FL/FR/RL/RR denote front-left, front-right, rear-left, and rear-right legs.\par
        \endgroup
    \end{minipage}
    \vspace{-1.8em}
\end{wraptable}

 serving target reachability. We quantify this effect in Table~\ref{tab:fl_posture_commands} using two indicators: the fault-leg load ratio $R_j=F_{z,j}/\sum_i F_{z,i}$, where $F_{z,i}$ denotes the vertical contact force of foot $i$ and $j$ denotes the faulty leg, and the fault-side tilt score $S_j=\max(0,\boldsymbol{d}_j^{\top}\boldsymbol{g}_{\mathrm{xy}})$, with $\boldsymbol{d}_j=\boldsymbol{r}_j^{\mathrm{xy}}/\|\boldsymbol{r}_j^{\mathrm{xy}}\|_2$. Here, $\boldsymbol{g}_{\mathrm{xy}}=[g_x,g_y]^{\top}$ is the horizontal projected-gravity vector, and $\boldsymbol{r}_j^{\mathrm{xy}}$ is the horizontal position vector from the base to the faulty foot $j$. Lower values indicate reduced loading and tilt toward the faulty FL leg. Compared with Robo-Duet and \textbf{w/o PAM}, our method lowers both metrics and yields safer reaching postures under different joint failure conditions while maintaining the commanded target.

\subsection{Simulation Experiments}
\label{subsec:sim_experiments}
    
To rigorously evaluate the system's robustness, we use $k_\tau$ to denote the actuator weakening level, with its detailed definition provided in Appendix~\ref{app:fault}. We then individually inject partial ($k_\tau = 0.1$) and complete ($k_\tau = 0.0$) weakening faults into the 12 lower-limb joints. Each evaluation episode sequentially performs locomotion tracking ($v_x = 0.4\,\mathrm{m/s}$) and continuous end-effector pose tracking. Evaluated over 1,000 randomized trials, the system is assessed by the \textbf{Survival Rate} (preventing mid-task falls) and \textbf{Workspace} (successfully reaching targets).

\begin{table*}[htbp]
\centering
\captionsetup{font=small, labelfont=bf, justification=justified}
\caption{Survival rate and workspace under partial ($k_\tau=0.1$) and complete ($k_\tau=0.0$) weakening across six representative joints. \textmd{RD denotes Robo-Duet, and ``-'' indicates immediate task failure.}}
\label{tab:sim_results_comprehensive}
\resizebox{1.0\textwidth}{!}{
\begin{tabular}{l cccc cccc cccc cccc}
\toprule
\multirow{3}{*}{\textbf{Faulty Joint}} & \multicolumn{8}{c}{\textbf{Survival Rate ($\%$) $\uparrow$}} & \multicolumn{8}{c}{\textbf{Workspace ($m^3$) $\uparrow$}} \\
\cmidrule(lr){2-9} \cmidrule(lr){10-17}
& \multicolumn{4}{c}{\textbf{Partial Weakening ($k_\tau = 0.1$)}} & \multicolumn{4}{c}{\textbf{Complete Weakening ($k_\tau = 0.0$)}} & \multicolumn{4}{c}{\textbf{Partial Weakening ($k_\tau = 0.1$)}} & \multicolumn{4}{c}{\textbf{Complete Weakening ($k_\tau = 0.0$)}} \\
\cmidrule(lr){2-5} \cmidrule(lr){6-9} \cmidrule(lr){10-13} \cmidrule(lr){14-17}
& RD & w/o FE & w/o PAM & \textbf{Ours} & RD & w/o FE & w/o PAM & \textbf{Ours} & RD & w/o FE & w/o PAM & \textbf{Ours} & RD & w/o FE & w/o PAM & \textbf{Ours} \\
\midrule
FL Calf   & -    & 55.2 & 81.4 & \textbf{84.5} & -    & 35.1 & 53.4 & \textbf{67.2} & -    & 0.29 & 0.54 & \textbf{0.59} & -    & 0.19 & 0.41 & \textbf{0.50} \\
RL Calf   & -    & 57.1 & 83.6 & \textbf{86.2} & -    & 37.4 & 55.6 & \textbf{68.8} & -    & 0.33 & 0.57 & \textbf{0.61} & -    & 0.21 & 0.43 & \textbf{0.52} \\
FR Thigh  & 52.3 & 68.4 & 91.2 & \textbf{93.4} & 39.4 & 55.2 & 73.1 & \textbf{84.3} & 0.33 & 0.45 & 0.75 & \textbf{0.77} & 0.24 & 0.35 & 0.63 & \textbf{0.70} \\
RL Thigh  & 55.1 & 71.3 & 93.6 & \textbf{94.8} & 42.1 & 58.7 & 76.3 & \textbf{86.5} & 0.35 & 0.48 & 0.77 & \textbf{0.79} & 0.27 & 0.38 & 0.66 & \textbf{0.72} \\
FL Hip    & 59.4 & 75.2 & 94.8 & \textbf{95.6} & 45.3 & 62.4 & 80.2 & \textbf{88.1} & 0.39 & 0.52 & 0.80 & \textbf{0.81} & 0.30 & 0.42 & 0.69 & \textbf{0.75} \\
RR Hip    & 63.2 & 79.1 & \textbf{96.5} & 96.2 & 50.2 & 66.3 & 83.4 & \textbf{89.8} & 0.43 & 0.56 & \textbf{0.84} & 0.82 & 0.33 & 0.45 & 0.71 & \textbf{0.77} \\
\midrule
\textbf{Average $\uparrow$} 
 & 37.5 & 67.7 & 90.2 & \textbf{91.8} & 29.4 & 52.5 & 70.3 & \textbf{80.8} & 0.25 & 0.44 & 0.71 & \textbf{0.73} & 0.19 & 0.34 & 0.59 & \textbf{0.66} \\
\bottomrule
\end{tabular}
}
\end{table*}

\textbf{Survival rate and workspace:} Table \ref{tab:sim_results_comprehensive} shows that FT-WBC consistently improves both survival rate and workspace over Robo-Duet across different joint failures. The results reveal a clear difficulty hierarchy, with calf and thigh faults causing larger performance drops because these joints strongly affect the base pitch required for reaching. Robo-Duet often fails immediately under critical calf faults due to the lack of fault awareness. The ablation results further show that both FE and PAM are necessary: without FE, the leg policy cannot perceive actuator damage and synthesize proper compensatory forces; without PAM, aggressive posture commands may increase workspace in low-risk cases but can destabilize the robot under severe faults. In contrast, full FT-WBC balances workspace and stability by adapting posture commands according to the estimated fault condition.

\begin{figure*}[htbp]
    \centering
    \includegraphics[width=1\linewidth]{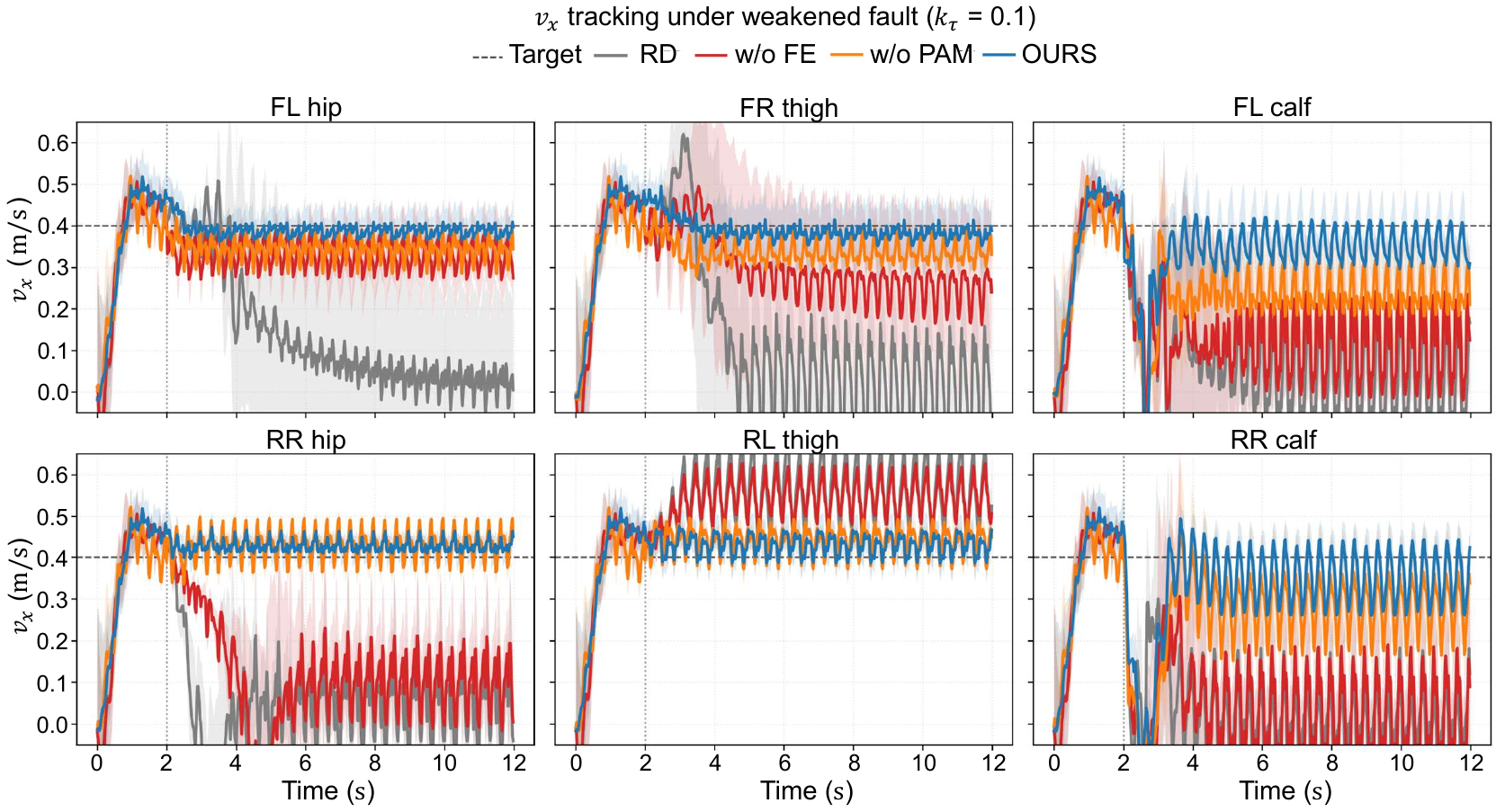}
    \caption{Forward linear velocity ($v_x$) tracking profiles under severe joint weakening ($k_\tau=0.1$). The target command (dashed grey) is set to $0.4\,\mathrm{m/s}$. The solid grey, red, orange, and blue lines represent Robo-Duet, w/o FE, w/o PAM, and our full framework, respectively. Our method significantly suppresses velocity oscillations and maintains tight tracking across diverse joint failures.}
    \label{fig:track_vx_6panel}
\end{figure*}

\textbf{Velocity tracking:} Figure~\ref{fig:track_vx_6panel} illustrates the real-time linear velocity ($v_x$) tracking profiles across six distinct failure scenarios. Compared with other methods, \textbf{RD} suffers from severe tracking collapse and a complete inability to maintain forward progress in multiple joint scenarios. Meanwhile, due to the inability to perceive actuator discrepancies, \textbf{w/o FE} exhibits divergent oscillations that quickly lead to falls. Incorporating the FE (as seen in \textbf{w/o PAM}) markedly improves survival compared to both \textbf{RD} and \textbf{w/o FE}, yet lacking posture adaptation still manifests as observable steady-state errors and speed fluctuations. Conversely, our full framework maintains precise tracking across all joint types. It rapidly synthesizes compensatory gaits to handle initial velocity drops caused by calf faults and effectively suppresses erratic swings induced by thigh faults. Overall, these results show the importance of combining rapid fault prediction with high-level posture adaptation.

\subsection{Real-World Deployment}
\label{subsec:real_world}

\begin{figure}[htbp]
    \centering
    \includegraphics[width=\linewidth]{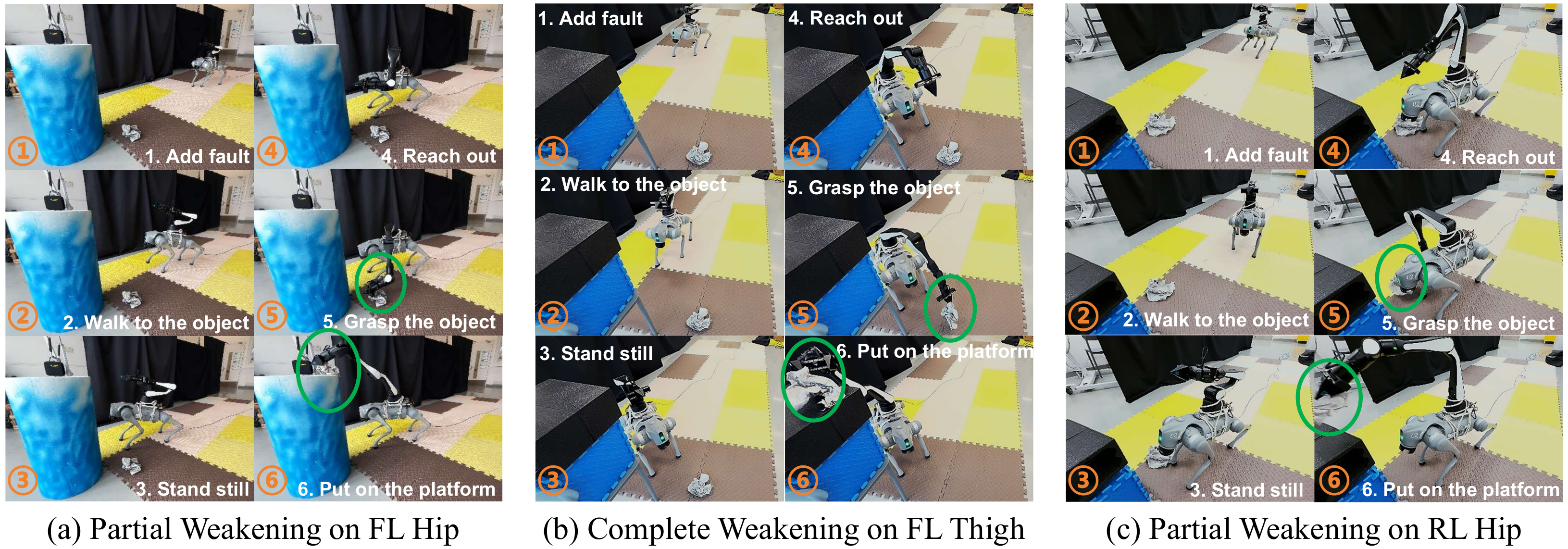}
    \caption{\justifying Real-world dynamic manipulation under different joint failures. Each subfigure shows fault-aware reaching on the left and the resulting pick-and-place motion on the right.}
    \label{fig:real_world_loco_wbc}
\end{figure}

To validate the zero-shot sim-to-real transferability and the practical robustness of our framework,  we design a ground-to-table pick-and-place task. The robot must pick up an object from the ground (requiring a deep downward pitch) and place it onto tables of varying heights: $50\,\mathrm{cm}$, $75\,\mathrm{cm}$, and $90\,\mathrm{cm}$. The $90\,\mathrm{cm}$ target is exceptionally demanding, forcing the quadruped into an extreme upward body pitch—a precarious maneuver when a support leg is compromised. As visually demonstrated in Figure \ref{fig:real_world_loco_wbc}, our framework enables the robot to safely execute these dynamic reaching and placement sequences despite being subjected to diverse and severe impairments, including partial weakening (FL Hip), complete weakening (FL Thigh), and an unmodeled locked fault (RL Calf).

\textbf{Physical trial results:} We conduct 20 physical trials for each height under three conditions: Partial Weakening (PW), Complete Weakening (CW), and an unmodeled Locked Fault (LF). Notably, evaluating against locked (jammed) joints explicitly verifies the framework's zero-shot generalization capability, as the policy is exclusively trained on torque-weakening faults. The quantitative results, measured by \textbf{Survival Rate} and \textbf{Success Rate} (completing the placement), are presented in Table \ref{tab:real_world_results}. As shown, the framework demonstrates remarkable resilience. For standard heights ($50\,\mathrm{cm}$ and $75\,\mathrm{cm}$), the system maintains high survival and success rates. 

\begin{wraptable}{r}{0.35\textwidth}
    \vspace{-1.55em}
    \centering
    \captionsetup{font=small, justification=centering}
    \caption{Pick-and-Place Performance.}
    \label{tab:real_world_results}
    
    \scriptsize 
    \setlength{\tabcolsep}{4pt}
    \begin{tabular}{cccc}
    \toprule
    \textbf{Height} & \textbf{Cond.} & \textbf{Survival} & \textbf{Success} \\
    \midrule
    \multirow{3}{*}{\boldmath\textbf{$50\,\mathrm{cm}$}} 
    & PW & 20/20 & 19/20\\
    & CW & 20/20 & 18/20\\
    & LF & 19/20 & 16/20 \\
    \midrule
    \multirow{3}{*}{\boldmath\textbf{$75\,\mathrm{cm}$}} 
    & PW & 20/20 & 18/20 \\
    & CW & 19/20 & 16/20 \\
    & LF & 17/20 & 14/20 \\
    \midrule
    \multirow{3}{*}{\boldmath\textbf{$90\,\mathrm{cm}$}} 
    & PW & 19/20 & 16/20\\
    & CW & 16/20 & 13/20\\
    & LF & 14/20 & 9/20 \\
    \bottomrule
    \end{tabular}
    \vspace{-0.8em}
\end{wraptable}

\textbf{Locked fault generalization:} At the extreme $90\,\mathrm{cm}$ height, our framework still guarantees a 70\% survival rate even under severe locked conditions, demonstrating strong adaptation to out-of-distribution actuator failures. While the task success rate drops to 45\%, this occurs because the PAM conservatively restricts the extreme upward pitch to prevent the CoM from escaping the degraded support polygon. This ``survival-first'' behavior aligns with the design philosophy of real-world fault-tolerant systems: prioritizing catastrophic-damage prevention over aggressive task completion.

\section{Conclusion}
\label{sec:conclusion}

This work presents FT-WBC, a fault-tolerant whole-body control framework that enables legged manipulators to maintain stable loco-manipulation under actuator failures. By incorporating fault estimation and posture adaptation into a decoupled upper- and lower-body policy architecture, FT-WBC dynamically adjusts base posture commands according to the current fault condition, allowing the robot to preserve whole-body stability while retaining as much arm workspace as possible.

Simulation and real-world experiments show that FT-WBC significantly improves survival rate and workspace under various joint weakening and complete failure scenarios, and further achieves effective zero-shot adaptation to locked faults that are not seen during training. These results demonstrate that the proposed framework improves not only locomotion stability under failures, but also the ability to sustain whole-body manipulation tasks. Overall, FT-WBC extends fault-tolerant control from conventional locomotion to loco-manipulation, providing a practical and scalable solution for reliable deployment of legged manipulators under actuator failures.


\section{Limitations}
\label{sec:Limitations}

Despite its significant progress, FT-WBC has two main limitations. First, it currently targets single-actuator failures and does not yet address concurrent multi-joint failures, where the possible fault combinations become high-dimensional. Second, our evaluations focus on stationary-base or sequential ``move-then-manipulate'' tasks, and extending fault-tolerant control to dynamic manipulation while walking remains future work for real-world legged manipulation.


\clearpage
\acknowledgments{If a paper is accepted, the final camera-ready version will (and probably should) include acknowledgments. All acknowledgments go at the end of the paper, including thanks to reviewers who gave useful comments, to colleagues who contributed to the ideas, and to funding agencies and corporate sponsors that provided financial support.}


\bibliography{example}  

\clearpage
\appendix
\section*{Appendix}

\section{Leg Policy and Arm Policy}
\label{app:leg_arm_policy}

\subsection{Policy Inputs and Actions}
\label{app:policy_inputs_actions}

FT-WBC adopts a decoupled upper- and lower-body policy architecture. The lower-body leg policy $\pi_{\mathrm{leg}}$ generates quadruped joint actions for locomotion stability and compensatory gait synthesis, while the upper-body arm policy $\pi_{\mathrm{arm}}$ tracks the target end-effector pose and produces both manipulator actions and a desired base posture plan. The FE and PAM provide the fault-aware connection between these two policies: FE supplies the predicted fault vector, and PAM adapts the base posture plan before it is executed by the lower-body controller.

The policy $\pi_{\mathrm{leg}}$ takes a 30-step lower-body observation history $\boldsymbol{o}_{t-29:t}^{\mathrm{leg}} \in \mathbb{R}^{1680}$ and target motion commands as input. The locomotion command $\boldsymbol{c}_t^{\mathrm{leg}} \in \mathbb{R}^{5}$ contains the desired forward linear velocity $v_x$, lateral velocity $v_y$, yaw angular velocity $\omega_z$, body pitch command, and body roll command, which specify the target motion and posture of the robot base. Each lower-body observation $\boldsymbol{o}_t^{\mathrm{leg}} \in \mathbb{R}^{56}$ contains the projected gravity $\boldsymbol{g}_t \in \mathbb{R}^{3}$, leg states composed of relative leg joint positions $\boldsymbol{q}_t^{\mathrm{leg}} - \boldsymbol{q}_0^{\mathrm{leg}} \in \mathbb{R}^{12}$ and leg joint velocities $\dot{\boldsymbol{q}}_t^{\mathrm{leg}} \in \mathbb{R}^{12}$, the previous leg action $\boldsymbol{a}_{t-1}^{\mathrm{leg}} \in \mathbb{R}^{12}$, lower-body velocity and posture commands $\boldsymbol{c}_t^{\mathrm{leg}} \in \mathbb{R}^{5}$, upper-body target commands $\boldsymbol{c}_t^{\mathrm{arm}} \in \mathbb{R}^{6}$, base states composed of the roll angle $\phi_t \in \mathbb{R}$ and pitch angle $\theta_t \in \mathbb{R}$, and the predicted fault vector $\hat{\boldsymbol{f}}_t \in \mathbb{R}^{12}$ from FE. $\pi_{\mathrm{leg}}$ outputs leg joint actions $\boldsymbol{a}_t^{\mathrm{leg}} \in \mathbb{R}^{12}$, which represent joint position offsets relative to the default joint positions. These actions are converted by the leg PD controller into joint torques $\boldsymbol{\tau}_t^{\mathrm{leg}}$, which are finally sent to the robot to drive the twelve leg joints.

$\pi_{\mathrm{arm}}$ tracks the target end-effector pose based on a 30-step arm observation history $\boldsymbol{o}_{t-29:t}^{\mathrm{arm}} \in \mathbb{R}^{600}$. The manipulation command $\boldsymbol{c}_t^{\mathrm{arm}} \in \mathbb{R}^{6}$ is expressed in a polar-style task parameterization: $l$ denotes the target distance from the base frame, while $p$ and $y$ define the target direction through pitch and yaw angles. The variables $\alpha$, $\beta$, and $\gamma$ further specify the desired roll, pitch, and yaw orientation of the end effector. Each arm observation $\boldsymbol{o}_t^{\mathrm{arm}} \in \mathbb{R}^{20}$ contains arm states composed of the relative joint positions of the six arm joints $\boldsymbol{q}_t^{\mathrm{arm}} - \boldsymbol{q}_0^{\mathrm{arm}} \in \mathbb{R}^{6}$, the previous arm action $\boldsymbol{a}_{t-1}^{\mathrm{arm}} \in \mathbb{R}^{6}$, the target end-effector command $\boldsymbol{c}_t^{\mathrm{arm}} \in \mathbb{R}^{6}$, and base states composed of the roll angle $\phi_t \in \mathbb{R}$ and pitch angle $\theta_t \in \mathbb{R}$. The output of $\pi_{\mathrm{arm}}$, $\boldsymbol{a}_{t}^{\mathrm{arm}} \in \mathbb{R}^{8}$, contains six arm joint actions and a desired base posture plan $\boldsymbol{u}_t \in \mathbb{R}^{2}$ sent to PAM. The six joint-action components are converted by the arm PD controller into arm joint torques $\boldsymbol{\tau}_t^{\mathrm{arm}}$, which are finally sent to the robot manipulator. This posture plan provides additional posture degrees of freedom for the arm and helps expand the reachable workspace of the end effector.

\subsection{Adaptation Modules}
\label{app:adaptation_modules}

Besides the FE and PAM, FT-WBC keeps the lightweight RMA-style adaptation modules used in the Robo-Duet policy decomposition. These modules should not be confused with the proposed fault-specific PAM: they do not output posture corrections and do not explicitly reason about the predicted fault vector $\hat{\boldsymbol{f}}_t$. Instead, they infer compact privileged context variables that are available in simulation during training but unavailable at deployment.

For each branch $b \in \{\mathrm{leg}, \mathrm{arm}\}$, where $\mathrm{leg}$ and $\mathrm{arm}$ denote the leg and arm branches, respectively, the adaptation module takes the corresponding observation history $\boldsymbol{o}_{t-H:t}^{b}$ as input and predicts a low-dimensional privileged vector $\hat{\boldsymbol{p}}_t^{b}=\phi_{\mathrm{ada}}^{b}(\boldsymbol{o}_{t-H:t}^{b})$. The notation $\boldsymbol{o}_{t-H:t}^{b}$ follows the Method section and denotes the history window from timestep $t-H$ to $t$. The network $\phi_{\mathrm{ada}}^{b}$ uses hidden dimensions $256$ and $128$, followed by a linear output layer whose dimension matches the privileged-observation dimension of that branch.

For the leg branch, the privileged vector is $\boldsymbol{p}_{t}^{\mathrm{leg}}=[\tilde{\mu}_{t},\tilde{e}_{t}]^{\top} \in \mathbb{R}^{2}$, where $\tilde{\mu}_{t}$ and $\tilde{e}_{t}$ denote the normalized ground-friction coefficient and restitution coefficient. For the arm branch, the privileged vector is $\boldsymbol{p}_{t}^{\mathrm{arm}}=[\tilde{\mu}_{t},\tilde{e}_{t},\boldsymbol{c}_{t}^{\mathrm{ee}},\boldsymbol{q}_{t}^{\mathrm{ee}}]^{\top} \in \mathbb{R}^{9}$, where $\boldsymbol{c}_{t}^{\mathrm{ee}}=[l_t,p_t,y_t]^{\top} \in \mathbb{R}^{3}$ is the LPY target representation in the base frame and $\boldsymbol{q}_{t}^{\mathrm{ee}}=[q_{w,t}^{\mathrm{ee}},q_{x,t}^{\mathrm{ee}},q_{y,t}^{\mathrm{ee}},q_{z,t}^{\mathrm{ee}}]^{\top} \in \mathbb{R}^{4}$ is the end-effector orientation quaternion relative to the base frame.

The adaptation module is trained with the MSE loss $\mathcal{L}_{\mathrm{ada}}^{b}=\|\hat{\boldsymbol{p}}_{t}^{b}-\boldsymbol{p}_{t}^{b}\|_{2}^{2}$ for $b \in \{\mathrm{leg}, \mathrm{arm}\}$, regressing the simulator privileged observations used by the teacher critic. During deployment, these privileged observations are removed, and the actor receives the predicted vector $\hat{\boldsymbol{p}}_t^{\mathrm{b}}$ instead. The leg actor concatenates the leg observation history $\boldsymbol{o}_{t-29:t}^{\mathrm{leg}} \in \mathbb{R}^{1680}$, the predicted privileged vector $\hat{\boldsymbol{p}}_t^{\mathrm{leg}} \in \mathbb{R}^{2}$, and the fault vector $\hat{\boldsymbol{f}}_t \in \mathbb{R}^{12}$ before the policy MLP, while the arm actor encodes the non-current part of the arm history $\boldsymbol{o}_{t-29:t-1}^{\mathrm{arm}} \in \mathbb{R}^{580}$ with a separate history encoder and combines it with $\boldsymbol{o}_t^{\mathrm{arm}} \in \mathbb{R}^{20}$ and $\hat{\boldsymbol{p}}_t^{\mathrm{arm}} \in \mathbb{R}^{9}$ before action generation. Thus, the adaptation modules provide compact environment and task context for robust policy execution, while explicit fault estimation and fault-aware base-posture modification are handled separately by FE and PAM.

The task command spaces during training and deployment are listed in Table~\ref{tab:command_ranges}. Here, $v_x$ and $\omega_z$ denote the base forward-velocity and yaw-rate commands, respectively; $l,p,y$ specify the end-effector target in the LPY representation; and $\alpha,\beta,\gamma$ specify the target end-effector orientation. The deployment ranges are slightly narrower for base velocity commands to improve real-world stability, while the manipulation command ranges are kept consistent with training.

\begin{table}[htbp]
    \centering
    \caption{Command Ranges for Training and Deployment}
    \label{tab:command_ranges}
    \small
    \setlength{\tabcolsep}{6pt}
    \begin{tabular}{lcc}
    \toprule
    \textbf{Parameters} & \textbf{Training} & \textbf{Deployment} \\
    \midrule
    $v_x$ ($\mathrm{m/s}$) & [-1.0, 1.0] & [-0.8, 0.8] \\
    $\omega_z$ ($\mathrm{rad/s}$) & [-1, 1] & [-0.8, 0.8] \\
    $l$ ($\mathrm{m}$) & [0.3, 0.77] & [0.3, 0.77] \\
    $p$ ($\mathrm{rad}$) & [-1.41, 1.41] & [-1.41, 1.41] \\
    $y$ ($\mathrm{rad}$) & [-1.57, 1.57] & [-1.57, 1.57] \\
    $\alpha$ ($\mathrm{rad}$) & [-1.41, 1.41] & [-1.41, 1.41] \\
    $\beta$ ($\mathrm{rad}$) & [-1.05, 1.05] & [-1.05, 1.05] \\
    $\gamma$ ($\mathrm{rad}$) & [-1.31, 1.31] & [-1.31, 1.31] \\
    \bottomrule
    \end{tabular}
\end{table}

\section{Reward Function Design}
\label{app:reward}

In the reward function, we do not encode prior knowledge about specific actuator damage models. Instead, fault-tolerant behavior is learned from randomized fault injection, history-based fault estimation, and fault-aware posture adaptation. The reward is designed to jointly encourage velocity tracking, stable base motion, smooth and energy-efficient joint behaviors, accurate end-effector tracking, and conservative compensatory behaviors under actuator failures. Training is divided into two stages. In \textbf{Stage I}, only the locomotion policy is optimized. In \textbf{Stage II}, the arm policy, PAM, and whole-body loco-manipulation rewards are activated after PPO iteration $T_{\mathrm{hyb}}$ ($8{,}000$ from scratch, or $2{,}000$ when resuming from a pretrained leg policy). After this switch, the per-timestep reward combines three components:
\begin{equation}
R_t = R_{\mathrm{locomotion}} + R_{\mathrm{manipulation}} + R_{\mathrm{fault}},
\end{equation}
where each component consists of weighted reward or penalty terms. All reward weights are multiplied by the simulation step $\Delta t=0.005\,\mathrm{s}$ at initialization.

\subsection{Locomotion Rewards}
\label{app:reward_loco}

The locomotion reward encourages the base to track commanded velocities while maintaining stable, smooth, and energy-efficient leg motions. Stage-I locomotion training also includes swing-foot clearance, contact-force and contact-velocity shaping, collision penalties, and foot-slip penalties. We use Raibert foot placement and contact shaping to regularize gait behaviors instead of explicitly enforcing a fixed pairwise gait pattern. The locomotion reward terms are summarized in Table~\ref{tab:reward_loco}.

\begin{table}[t]
\centering
\caption{Locomotion reward terms in Stage I.}
\label{tab:reward_loco}
\scriptsize
\setlength{\tabcolsep}{4pt} 
\renewcommand{\arraystretch}{1.25} 

\resizebox{\textwidth}{!}{%
\begin{tabular}{>{\centering\arraybackslash}p{0.16\textwidth}>{\centering\arraybackslash}p{0.25\textwidth}>{\raggedright\arraybackslash}p{0.46\textwidth}c}
\toprule
\multicolumn{1}{c}{Reward name} & \multicolumn{1}{c}{Reward formula} & \multicolumn{1}{c}{Function and symbols} & \multicolumn{1}{c}{Weight} \\
\midrule
$R_{\mathrm{tracking\_lin}}$ & $\exp(-\|\boldsymbol{c}_{xy}-\boldsymbol{v}_{xy}\|^2/\sigma_{\mathrm{tracking}}^2)$ & Tracks commanded planar velocity; $\boldsymbol{c}_{xy}$ and $\boldsymbol{v}_{xy}$ are commanded and actual base velocities, with $\sigma_{\mathrm{tracking}}=0.25$. & $1.0$ \\

$R_{\mathrm{tracking\_ang}}$ & $\exp(-({\omega_z^{\mathrm{cmd}}-\omega_z})^2/\sigma_{\mathrm{yaw}})$ & Tracks commanded yaw rate and reduces heading drift; $\omega_z^{\mathrm{cmd}}$ and $\omega_z$ are commanded and actual yaw rates. & $0.5$ \\

$R_{\mathrm{lin\_vel\_}z}$ & $v_z^2$ & Penalizes vertical base motion and suppresses height oscillation; $v_z$ is the base velocity along the vertical axis. & ${-2.0\times10^{-2}}$ \\

$R_{\mathrm{ang\_vel\_}xy}$ & $\omega_x^2+\omega_y^2$ & Penalizes roll/pitch angular velocities to stabilize the base and reduce body oscillation during locomotion. & ${-1.0\times10^{-3}}$ \\

$R_{\mathrm{torque}}$ & $\sum_j \tau_j^2$ & Reduces large joint torques; $\tau_j$ is the torque applied to joint $j$. & ${-1.0\times10^{-4}}$ \\

$R_{\mathrm{dof\_vel}}$ & $\sum_j \dot{q}_{j,t}^{2}$ & Penalizes fast joint motion; $\dot{q}_{j,t}$ denotes \mbox{joint velocity} at \mbox{timestep $t$.} & ${-1.0\times10^{-4}}$ \\

$R_{\mathrm{dof\_acc}}$ & $\sum_j ((\dot{q}_{j,t}-\dot{q}_{j,t-1})/\Delta t)^2$ & Smooths joint trajectories by penalizing abrupt acceleration changes; $\Delta t$ is the simulation step. & ${-2.5\times10^{-7}}$ \\

$R_{\mathrm{action\_rate}}$ & $\sum_j(a_{j,t}-a_{j,t-1})^2$ & Encourages smooth actions; $a_{j,t}$ is the current action of joint $j$. & ${-1.0\times10^{-2}}$ \\

$R_{\mathrm{loco\_energy}}$ & $\sum_{j\in\mathrm{legs}}(\tau_j\dot{q}_j)^2$ & Penalizes leg-joint power to encourage \mbox{efficient compensatory gaits.} & ${-4.0\times10^{-5}}$ \\

$R_{\mathrm{smooth}}$ & $\sum_i m_{t,i}(q^{\mathrm{target}}_{t,i}-2q^{\mathrm{target}}_{t-1,i}+q^{\mathrm{target}}_{t-2,i})^2$ & Penalizes target-position jerk; $m_{t,i}$ masks inactive actions. & ${-1.0\times10^{-1}}$ \\

$R_{\mathrm{raibert}}$ & $\sum_{i=1}^{4}\|\boldsymbol{f}_{i}^{\mathrm{des}}-\boldsymbol{f}_{i}^{\mathrm{act}}\|^2$ & Regularizes foot placement for stable stepping; $\boldsymbol{f}_{i}^{\mathrm{des}}$ and $\boldsymbol{f}_{i}^{\mathrm{act}}$ are desired and actual foot positions. & $-10.0$ \\

$R_{\mathrm{clearance}}$ & $\sum_i(h_i^{\mathrm{des}}-h_i)^2$ & Shapes swing-foot clearance to avoid low swings; $h_i^{\mathrm{des}}$ and $h_i$ are desired and actual foot heights. & $-30.0$ \\

$R_{\mathrm{contact\_force}}$ & $\sum_i\max(\|\boldsymbol{F}_i\|-F_{\max},0)^2$ & Regularizes contact forces; $\boldsymbol{F}_i$ is foot force \mbox{with threshold $F_{\max}$.} & $4.0$ \\

$R_{\mathrm{contact\_vel}}$ & $\sum_i \mathcal{I}_i^{c}\|\boldsymbol{v}_i^{\mathrm{foot}}\|^2$ & Penalizes stance-foot motion; $\mathcal{I}_i^{c}$ indicates contact of foot $i$. & $4.0$ \\

$R_{\mathrm{collision}}$ & $\sum_k \mathcal{I}_k^{\mathrm{collision}}$ & Penalizes undesired contacts or collisions; $\mathcal{I}_k^{\mathrm{collision}}$ indicates collision at body part $k$. & $-5.0$ \\

$R_{\mathrm{ori\_ctrl}}$ & $\|\boldsymbol{g}_{xy}^{\mathrm{proj}}-\boldsymbol{g}_{xy}^{\mathrm{des}}\|^2$ & Tracks desired base posture for stable balance; $\boldsymbol{g}_{xy}^{\mathrm{proj}}$ and $\boldsymbol{g}_{xy}^{\mathrm{des}}$ are actual and desired gravity projections. & $-5.0$ \\

$R_{\mathrm{slip}}$ & $\sum_i\mathcal{I}_i^{c}\|\boldsymbol{v}_{i,xy}^{\mathrm{foot}}\|^2$ & Penalizes stance-foot slip; $\boldsymbol{v}_{i,xy}^{\mathrm{foot}}$ is horizontal foot velocity. & ${-4.0\times10^{-2}}$ \\
\bottomrule
\end{tabular}
}
\end{table}

\subsection{Manipulation Rewards}
\label{app:reward_manip}

In Stage II, the arm policy is activated and outputs both six arm-joint actions and a two-dimensional base pitch/roll posture plan. The manipulation rewards encourage the end effector to track the commanded pose while keeping the base posture plan smooth and physically feasible. During Stage II, the linear and angular velocity tracking weights are reduced to $0.7$ and $0.25$, respectively, because the robot must balance velocity tracking with end-effector reachability and posture adaptation. The orientation-control penalty is increased to strengthen base posture regulation during reaching. The Stage-II reward terms are shown in Table~\ref{tab:reward_manip}.

\begin{table}[t]
\centering
\caption{Manipulation and updated locomotion reward terms in Stage II.}
\label{tab:reward_manip}
\scriptsize
\setlength{\tabcolsep}{4pt} 
\renewcommand{\arraystretch}{1.3} 

\resizebox{\textwidth}{!}{%
\begin{tabular}{>{\centering\arraybackslash}p{0.16\textwidth}>{\centering\arraybackslash}p{0.25\textwidth}>{\raggedright\arraybackslash}p{0.46\textwidth}c}
\toprule
\multicolumn{1}{c}{Reward name} & \multicolumn{1}{c}{Reward formula} & \multicolumn{1}{c}{Function and symbols} & \multicolumn{1}{c}{Weight} \\
\midrule
$R_{\mathrm{tracking\_lin}}$ & $\exp(-\|\boldsymbol{c}_{xy}-\boldsymbol{v}_{xy}\|^2/\sigma_{\mathrm{tracking}}^2)$ & Maintains base translation tracking during reaching and posture adaptation; symbols follow Table~\ref{tab:reward_loco}. & $0.7$ \\

$R_{\mathrm{tracking\_ang}}$ & $\exp(-({\omega_z^{\mathrm{cmd}}-\omega_z})^2/\sigma_{\mathrm{yaw}})$ & Maintains yaw-rate tracking without over-constraining the adaptive base posture; symbols follow Table~\ref{tab:reward_loco}. & $0.25$ \\

$R_{\mathrm{manip}}$ & $\exp(-w_{\mathrm{lpy}}e_{\mathrm{lpy}}-w_{\mathrm{rpy}}e_{\mathrm{rpy}})$ & Tracks commanded end-effector pose; $e_{\mathrm{lpy}}$ and $e_{\mathrm{rpy}}$ denote position and orientation errors. & $1.0$ \\

$R_{\mathrm{ori\_ctrl}}$ & $\|\boldsymbol{g}_{xy}^{\mathrm{proj}}-\boldsymbol{g}_{xy}^{\mathrm{des}}\|^2$ & Strengthens base posture tracking during reaching and whole-body balancing; symbols follow Table~\ref{tab:reward_loco}. & $-10.0$ \\

$R_{\mathrm{ori\_heur}}$ & $\|\boldsymbol{g}_{xy}^{\mathrm{proj}}-\boldsymbol{g}_{xy}^{\mathrm{safe}}\|^2$ & Penalizes unsafe body tilting with safe posture reference $\boldsymbol{g}_{xy}^{\mathrm{safe}}$. & $-2.0$ \\

$R_{\mathrm{hip\_act}}$ & $\sum_{j\in\mathrm{hip}} a_{j,t}^2$ & Regularizes hip actions to suppress aggressive lateral posture changes during reaching task. & ${-5.0\times10^{-2}}$ \\

$R_{\mathrm{plan\_smooth}}$ & $\|\boldsymbol{u}_t-\boldsymbol{u}_{t-1}\|^2$ & Smooths base posture plans to prevent high-frequency oscillations, where $\boldsymbol{u}_t$ represents the arm-policy pitch/roll plan. & ${-1.0\times10^{-1}}$ \\

$R_{\mathrm{plan\_limit}}$ & $\|\max(|\boldsymbol{u}_t|-\boldsymbol{u}_{\max},0)\|^2$ & Penalizes kinematically infeasible pitch/roll plans that strictly exceed the predefined posture-plan bound $\boldsymbol{u}_{\max}$. & $-5.0$ \\

$R_{\mathrm{arm\_energy}}$ & $\sum_{j\in\mathrm{arm}}(\tau_j\dot{q}_j)^2$ & Penalizes excessive arm-joint power to reduce actuator wear and ensure stable, smooth manipulation. & ${-4.0\times10^{-5}}$ \\
\bottomrule
\end{tabular}
}
\end{table}

\subsection{Fault-Oriented Rewards}
\label{app:reward_fault}

Fault-oriented rewards are used to encourage compensatory behaviors after actuator degradation. These terms are not designed from a detailed fault dynamics model; instead, they provide weak guidance that helps the policy reduce the reliance on failed joints and redistribute support to healthier legs. Let $F_j\in\{0,1,2\}$ denote the fault label of leg joint $j$, where $0$ indicates a healthy joint, $1$ indicates a locked joint, and $2$ indicates a weakened joint.

\begin{itemize}
    \item \textbf{Reward for hold faulty-joint motion:}
    \begin{equation}
    r_{\mathrm{motion}}^{\mathrm{fault}} = \sum_{j=1}^{12} m_j^{\mathrm{fault}}\dot{q}_j^2,
    \end{equation}
    where $m_j^{\mathrm{fault}}$ is a binary fault mask, with $m_j^{\mathrm{fault}}=1$ if joint $j$ is faulty ($F_j>0$) and $m_j^{\mathrm{fault}}=0$ otherwise. This penalty suppresses high velocities on impaired actuators, encouraging the policy to reduce failed-joint participation and avoid unstable commands that drive damaged motors.

    \item \textbf{Reward of contralateral compensatory support:}
    \begin{equation}
    r_{\mathrm{axis}} = \exp\!\left(-\frac{y_{\mathrm{healthy}}^2}{\sigma^2}\right), \quad \sigma=0.2\,\mathrm{m},
    \end{equation}
    where $y_{\mathrm{healthy}}$ is the lateral coordinate of the healthy contralateral foot in the yaw-aligned frame. When one leg in a front or rear pair is faulty, this reward encourages the healthy contralateral leg to move closer to the body sagittal plane, creating a more effective support configuration. Both the front pair and rear pair can contribute, so $r_{\mathrm{axis}}\in[0,2]$.
\end{itemize}

The final fault-oriented reward is
\begin{equation}
R_{\mathrm{fault}} = w_m r_{\mathrm{motion}}^{\mathrm{fault}} + w_y r_{\mathrm{axis}},
\end{equation}
where $w_m$ and $w_y$ are the corresponding weights. The values are listed in Table~\ref{tab:reward_fault}.

\begin{table}[t]
\centering
\caption{Fault-oriented reward weights.}
\label{tab:reward_fault}
\small
\begin{tabular}{llc}
\toprule
Category & Reward & Weight \\
\midrule
Hold faulty-joint motion & $r_{\mathrm{motion}}^{\mathrm{fault}}$ & $-0.2$ \\
Contralateral support & $r_{\mathrm{axis}}$ & $+0.6$ \\
\bottomrule
\end{tabular}
\end{table}

\section{Fault Definition and Fault Curriculum}
\label{app:fault}

Although policy training primarily uses torque weakening, zero-shot experiments also inject locked joints; both modes are defined below. Each leg joint $j$ is labeled by $F_j\in\{0,1,2\}$, corresponding to normal, locked, and weakened states. Consistent with the main text, the weakening coefficient $k_\tau\in[0,1]$ scales available joint torque: $k_\tau=1$ is healthy, $k_\tau=0.1$ denotes \textit{partial weakening}, and $k_\tau=0.0$ denotes \textit{complete weakening} at evaluation time.

\textbf{Locked joint ($F_j=1$):}
\begin{equation}
q_j^{\mathrm{cmd}\prime} = \mathrm{clip}\!\left(q_j^{\mathrm{cmd}},\, c_j - q_{\mathrm{thr}},\, c_j + q_{\mathrm{thr}}\right), \quad q_{\mathrm{thr}}=0.05\,\mathrm{rad}.
\end{equation}

\textbf{Weakened motor ($F_j=2$):}
\begin{equation}
\tau_j' = k_{\tau,j}\,\tau_j, \quad k_{\tau,j}\in[0,1].
\end{equation}

During training, a faulty episode impairs all three joints of one leg (FL, FR, RL, or RR) with probability $0.95$, enabling the policy to repeatedly experience leg-level actuator degradation and learn compensatory support behaviors across different impaired support configurations.

\textbf{Weakening-severity curriculum.} Uniform sampling of $k_{\tau,j}$ yields too few near-complete weakening samples. To emphasize severe actuator degradation, we linearly increase the fraction of samples whose $k_{\tau,j}$ approaches complete weakening ($k_\tau=0$):
\begin{equation}
\rho(t) = \rho_0 + \mathrm{clip}\!\left(\frac{t-t_0}{T}, 0, 1\right)(\rho_1-\rho_0),
\end{equation}
with $\rho_0=0$, $\rho_1=0.3$, $t_0=0$, and $T=5000$. While $k_{\tau,j}\sim\mathcal{U}(0, 0.25)$, the fraction $\rho(t)$ is stratified toward the near-complete regime, producing a smooth transition from mild weakening to near-complete torque loss during training.

\section{Hyperparameters of Training}
\label{app:hyperparams}

\subsection{Adaptation Module and PPO}
\label{app:adaptation}

The adaptation modules are described in Appendix~\ref{app:adaptation_modules}. In Table~\ref{tab:hyperparams_ppo}, $d_{\mathrm{hist}}^{\mathrm{b}}$ denotes the flattened observation-history dimension of branch $\mathrm{b}\in\{\mathrm{l},\mathrm{a}\}$, $d_{\mathrm{obs}}^{\mathrm{a}}$ denotes the current arm-observation dimension, and $d_{\mathrm{priv}}^{\mathrm{b}}$ denotes the privileged-vector dimension predicted by the corresponding adaptation module. In our implementation, the leg branch uses $d_{\mathrm{obs}}^{\mathrm{l}}=56$, $d_{\mathrm{hist}}^{\mathrm{l}}=56\times30=1680$, and $d_{\mathrm{priv}}^{\mathrm{l}}=2$, while the arm branch uses $d_{\mathrm{obs}}^{\mathrm{a}}=20$, $d_{\mathrm{hist}}^{\mathrm{a}}=20\times30=600$, and $d_{\mathrm{priv}}^{\mathrm{a}}=9$. Their implementation details and the PPO training hyperparameters are summarized below.
\begin{table}[H]
\centering
\caption{Hyperparameters for the adaptation module and PPO.}
\label{tab:hyperparams_ppo}
\small
\begin{tabular}{lc}
\toprule
Hyperparameter & Value \\
\midrule
Leg adaptation MLP & $d_{\mathrm{hist}}^{l} \to 256 \to 128 \to d_{\mathrm{priv}}^{l}$ \\
Arm adaptation MLP & $d_{\mathrm{hist}}^{a} \to 256 \to 128 \to d_{\mathrm{priv}}^{a}$ \\
Adaptation target & branch privileged observations \\
Adaptation loss & MSE \\
Arm history encoder & $(d_{\mathrm{hist}}^{a}-d_{\mathrm{obs}}^{a}) \to 512 \to 256 \to 128$ \\
Adaptation module learning rate & $5\times10^{-4}$ \\
Number of learning epochs & $5$ \\
Number of mini-batches & $4$ \\
Steps per environment per iteration & $24$ \\
PPO learning rate & $5\times10^{-4}$ \\
Discount factor $\gamma$ & $0.99$ \\
GAE parameter $\lambda$ & $0.95$ \\
PPO clip $\epsilon$ & $0.2$ \\
Entropy coefficient & $0.01$ \\
Value loss coefficient & $1.0$ \\
Desired KL divergence & $0.01$ \\
Maximum gradient norm & $1.0$ \\
Default total iterations & $10^5$ \\
Parallel environments & $4096$ \\
Simulation step $\Delta t$ & $0.005$\,s \\
Control decimation & $4$ \\
Episode length & $20$\,s \\
Policy observation history & $30$ steps \\
Hybrid switch iteration $T_{\mathrm{hyb}}$ & $10{,}000$ ($2{,}000$ if resuming legs) \\
Fault probability per episode & $0.95$ \\
Actor--critic hidden layers & $[512, 256, 128]$, ELU \\
\bottomrule
\end{tabular}
\end{table}

\subsection{Fault Estimator}
\label{app:fe}

Training hyperparameters of the FE (Sec.~\ref{subsec:fault_estimator}) are summarized in Table~\ref{tab:hyperparams_fe}.

\begin{table}[H]
\centering
\caption{Fault Estimator hyperparameters.}
\label{tab:hyperparams_fe}
\small
\begin{tabular}{lc}
\toprule
Hyperparameter & Value \\
\midrule
Input dimension & $5\times56=280$ \\
History length $K$ & $5$ \\
Learning rate & $1\times10^{-3}$ \\
Loss & MSE on fault probability \\
Network & $280 \to 512 \to 256 \to 128 \to 12$ \\
GT-fault warmup iterations & $3000$ \\
\bottomrule
\end{tabular}
\end{table}

\subsection{Posture Adaptation Module}
\label{app:pam}

Implementation and training hyperparameters of the PAM (Sec.~\ref{subsec:posture_adaptation}) are summarized in Table~\ref{tab:hyperparams_pam}.

\begin{table}[H]
\centering
\caption{PAM structure and parameters.}
\label{tab:hyperparams_pam}
\small
\begin{tabular}{lc}
\toprule
Parameter & Value \\
\midrule
Input dimension & $2+12=14$ (plan + fault) \\
Hidden layers & $128 \to 64$ \\
Activation & ELU \\
Output dimension & $2$ (pitch / roll) \\
Output activation & $\tanh$ \\
Training & Joint PPO with arm policy \\
GT-fault warmup & $3000$ iterations \\
Pitch clip range (training) & $[-0.4,\, 0.3]$\,rad \\
Roll clip range (training) & $[-0.4,\, 0.4]$\,rad \\
\bottomrule
\end{tabular}
\end{table}

\section{Additional Locked-Fault Experiments}
\label{app:locked_fault_experiments}

We further evaluate zero-shot robustness under locked actuator faults, which are not included in the torque-weakening training distribution. Following the simulation protocol in Sec.~\ref{subsec:sim_experiments}, each episode evaluates locomotion tracking with $v_x=0.4\,\mathrm{m/s}$ and continuous end-effector pose tracking. Table~\ref{tab:locked_fault_results} reports the survival rate, reachable workspace, and velocity-tracking error under locked failures. The velocity-tracking error is computed along the forward $x$ direction.

\begin{table*}[htbp]
\centering
\captionsetup{font=small, labelfont=bf, justification=justified}
\caption{Survival rate, workspace, and velocity-tracking error under locked faults. RD denotes Robo-Duet, and ``-'' indicates immediate task failure. The velocity-tracking error is measured along the forward $x$ direction.}
\label{tab:locked_fault_results}
\resizebox{1.0\textwidth}{!}{

\begin{tabular}{c cccc cccc cccc}
\toprule 
\multirow{2}{*}{\textbf{Locked Joint}} & \multicolumn{4}{c}{\textbf{Survival Rate ($\%$) $\uparrow$}} & \multicolumn{4}{c}{\textbf{Workspace ($\mathrm{m^3}$) $\uparrow$}} & \multicolumn{4}{c}{\textbf{Vel. Tracking Err. ($\mathrm{m/s}$) $\downarrow$}} \\
\cmidrule(lr){2-5} \cmidrule(lr){6-9} \cmidrule(lr){10-13}
& RD & w/o FE & w/o PAM & \textbf{Ours} & RD & w/o FE & w/o PAM & \textbf{Ours} & RD & w/o FE & w/o PAM & \textbf{Ours} \\
\midrule

FL Calf  & $32.4\%$ & $64.3\%$ & $78.2\%$ & $\mathbf{84.1\%}$ & $0.125$ & $0.254$ & $0.481$ & $\mathbf{0.556}$ & $0.395$ & $0.382$ & $0.186$ & $\mathbf{0.042}$ \\
RR Calf  & $35.2\%$ & $66.5\%$ & $79.5\%$ & $\mathbf{85.3\%}$ & $0.142$ & $0.278$ & $0.505$ & $\mathbf{0.578}$ & $0.392$ & $0.375$ & $0.174$ & $\mathbf{0.038}$ \\
FR Thigh & $60.1\%$ & $76.2\%$ & $88.4\%$ & $\mathbf{91.2\%}$ & $0.324$ & $0.426$ & $0.641$ & $\mathbf{0.691}$ & $0.321$ & $0.298$ & $0.082$ & $\mathbf{0.015}$ \\
RL Thigh & $64.8\%$ & $80.4\%$ & $90.2\%$ & $\mathbf{93.1\%}$ & $0.346$ & $0.452$ & $0.662$ & $\mathbf{0.714}$ & $0.148$ & $0.136$ & $0.024$ & $\mathbf{0.012}$ \\
FL Hip   & $73.5\%$ & $84.1\%$ & $92.6\%$ & $\mathbf{96.2\%}$ & $0.371$ & $0.483$ & $0.694$ & $\mathbf{0.732}$ & $0.344$ & $0.312$ & $0.051$ & $\mathbf{0.018}$ \\
RR Hip   & $71.2\%$ & $85.3\%$ & $93.1\%$ & $\mathbf{96.5\%}$ & $0.389$ & $0.506$ & $0.711$ & $\mathbf{0.742}$ & $0.388$ & $0.364$ & $0.032$ & $\mathbf{0.021}$ \\
\midrule
\textbf{Average} & $56.2\%$ & $76.1\%$ & $87.0\%$ & $\mathbf{91.1\%}$ & $0.283$ & $0.400$ & $0.616$ & $\mathbf{0.669}$ & $0.331$ & $0.311$ & $0.092$ & $\mathbf{0.024}$ \\
\bottomrule
\end{tabular}
}
\end{table*}

\textbf{Locked fault generalization:} Table~\ref{tab:locked_fault_results} shows that FT-WBC generalizes well from torque weakening to unseen locked joints. Across six locked-fault cases, our method obtains the best average survival rate ($91.1\%$), workspace ($0.669\,\mathrm{m}^2$), and velocity-tracking error ($0.024\,\mathrm{m/s}$). Locked calf faults remain the most challenging because they directly affect stance support and foot clearance, forcing the base to undergo significant vertical disturbances. The clear degradation in the w/o FE variant highlights that without an explicit fault belief, the leg policy exhibits severe trajectory tracking lag. Compared with Robo-Duet and the two ablations, full FT-WBC better balances survival, workspace, and tracking accuracy, confirming the benefit of combining explicit fault prediction with posture adaptation under jammed actuators. This dual-loop integration effectively prevents abrupt actuator locking from inducing catastrophic full-body over-turning.

\clearpage %

\section{Snapshots of Simulation Experiments}
\label{app:snapshots_experiments}

This section supports the evaluation of FT-WBC using simulation snapshots of velocity tracking and workspace tasks under PW, CW, and LF conditions. These sequences highlight the coordination between full-body posture adaptation and dynamic gait compensation under joint failures.

\begin{figure*}[htbp]
    \centering
    \includegraphics[width=1.0\linewidth]{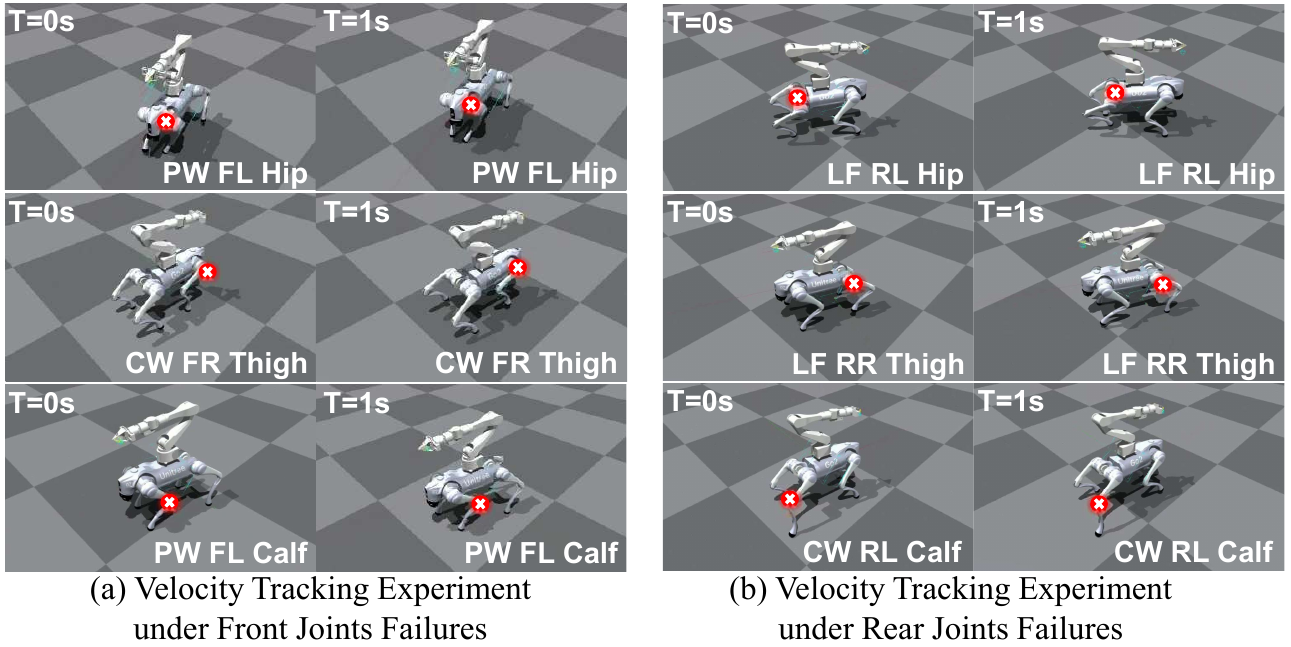} %
    \hfill
    
    \centering
    \includegraphics[width=1.0\linewidth]{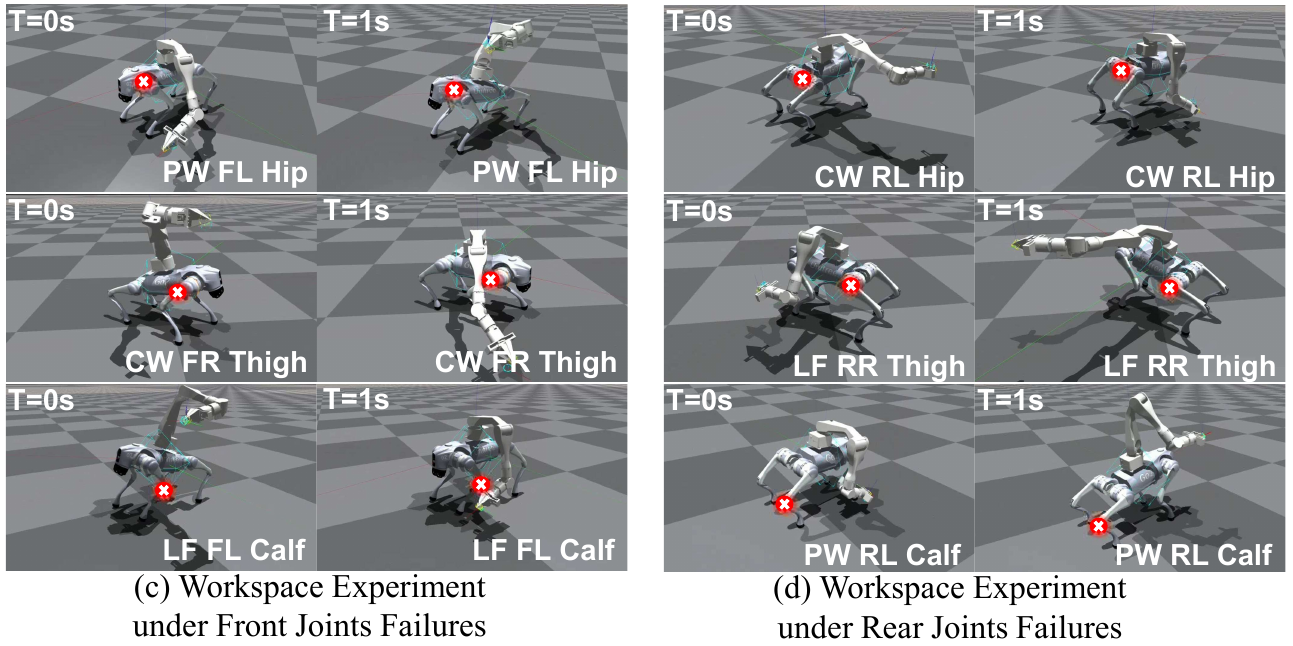} 
    \hfill
   
    \caption{Sequential simulation snapshots at $T=0\,\mathrm{s}$ and $T=1\,\mathrm{s}$ under representative PW, CW, and LF joint failure configurations for velocity tracking (top) and workspace test (bottom). Red crosses indicate the joints with injected failures across the quadruped platform.}
    \label{fig:simulation_snapshots}
\end{figure*}

As shown in Figure~\ref{fig:simulation_snapshots} (top rows), during locomotion ($v_x = 0.4\,\mathrm{m/s}$), the FE continuously estimates actuator failures to help synthesize compensatory gaits. The snapshots at $T=0\,\mathrm{s}$ and $T=1\,\mathrm{s}$ exhibit adapted walking postures under hip, thigh, and calf failures, where healthy limbs dynamically adjust touchdown behaviors and swing heights to sustain target velocity.

Furthermore, Figure~\ref{fig:simulation_snapshots} (bottom rows) highlights the robust whole-body control capabilities of our strategy under unexpected joint failures. The adaptive modulation effectively prevents the CoM from escaping the safe support polygon, thereby simultaneously preserving the necessary end-effector workspace and enhancing the overall closed-loop tolerance against actuator failures.

\end{document}